\pdfoutput=1
\documentclass[11pt,a4paper]{article}
\usepackage{emnlp2021}
\usepackage{times}
\usepackage{latexsym}

\usepackage{microtype}
\usepackage{multirow}
\usepackage{graphicx}
\usepackage{adjustbox}
\usepackage{comment}
\usepackage{tikz}
\usepackage[T1]{fontenc}
\usepackage[utf8]{inputenc}
\usepackage{enumitem}
\usepackage{booktabs}

\usepackage{cmap}
\usepackage[utf8]{inputenc}
\usepackage[LAE, T1]{fontenc}
\usepackage[farsi, english]{babel}

\title{\textsc{Sina-BERT}: A Pre-trained Language Model for Analysis of Medical Texts in Persian}

\author{
	Nasrin Taghizadeh \\
  Smart Persian Medical Assistant Group \\  \footnotesize{\FR{(هوش افزار سلامت)}} \\
  \texttt{nsr.taghizadeh@ut.ac.ir} \\
  \And
  Ehsan Doostmohammadi \\
  Sharif University of Technology \\
  \texttt{e.doostm72@student.sharif.edu} \\
  \AND
  Elham Seifossadat \\
  Sharif University of Technology \\
  \texttt{seifossadat@ce.sharif.edu} \\
  \And
  Hamid R. Rabiee \\
  Sharif University of Technology \\
  \texttt{rabiee@sharif.edu} \\
  \AND
  Maedeh S. Tahaei \\
  Sharif University of Technology \\
  Smart Persian Medical Assistant Group \footnotesize{\FR{(هوش افزار سلامت)}}  \\
  \texttt{mstahaei@aut.ac.ir} \\
 }

\date{}

\begin{document}
\maketitle

\begin{abstract}

We have released \textsc{Sina-BERT}, a language model pre-trained on BERT \cite{devlin2018bert} to address the lack of a high-quality Persian language model in the medical domain. \textsc{Sina-BERT} utilizes pre-training on a large-scale corpus of medical contents including formal and informal texts collected from a variety of online resources in order to improve the performance on health-care related tasks. We employ \textsc{Sina-BERT} to complete following representative tasks: categorization of medical questions, medical sentiment analysis, and medical question retrieval. For each task, we have developed Persian annotated data sets for training and evaluation and learnt a representation for the data of each task especially complex and long medical questions. With the same architecture being used across tasks, \textsc{Sina-BERT} outperforms BERT-based models that were previously made available in the Persian language.

\end{abstract}

\section{Introduction}

Patients, physicians and healthcare professionals are generating textual information every day using diverse formats that can be found in online resources. To improve the diagnosis and treatment of disease, text mining techniques are becoming increasingly important. Developing computational models of disease and applying these models to massive collections of textual information are two major challenges of computational medicine \cite{rakocevic2013computational}.

Text mining methods have been considered in multiple research studies in medicine; the most important ones being Named Entity Recognition (NER), personal data anonymization, knowledge discovery \cite{bokharaeian2017snpphena}, and terminology extraction \cite{luque2019advanced}. By employing text mining techniques, several healthcare systems can be developed such as Question Answering Systems \cite{ozyurt2020bio} and medical specialized search engines \cite{luo2008medsearch}.

Recent progress in medical text mining is due to advancements in the deep learning techniques used for natural language processing (NLP). In particular, language models have shown great advances in most NLP tasks and many current state-of-the-art methods often rely on Transformer-based pre-trained language models \cite{devlin2018bert,radford2018improving}.

While there are several BERT-based language models for the medical domain in English \cite{lee2020biobert,beltagy2019scibert,rasmy2020med}, Persian lacks such resources. In this paper, we present \textsc{Sina-BERT}, which is a pre-trained language representation model for the Persian biomedical domain. First, we initialize \textsc{Sina-BERT} with weights from ParsBERT \cite{farahani2020parsbert}, which is a general domain Persian language model. Then, \textsc{Sina-BERT} is pre-trained on large Persian medical corpora, which are collected from medical and health related websites, journals, books, forums and news websites. These corpora contain 2.8M documents from both formal and informal texts.

To show the effectiveness of our language model in medical text mining, \textsc{Sina-BERT} has been fine-tuned and evaluated on the following popular medical text mining tasks: question classification, sentiment analysis, and question retrieval. We also provide an annotated data set for each task and compare the performance of \textsc{Sina-BERT} against the state-of-the-art models. Therefore, the contributions of this paper can be listed as follows:
\begin{itemize}[noitemsep,nolistsep]
    \item A large scale Persian medical corpus. 
    \item A pre-trained language model for the Persian medical domain.
    \item A database containing 200k Persian medical questions answered by professional physicians for the task of question retrieval. 
    \item Annotated data sets for tasks of medical sentiment analysis and categorizing medical questions in Persian.
    \item Three data sets for automatic evaluation of medical retrieval systems in Persian.
    \item Learning a representation for medical complex and long questions based on deep sentence representation and ranking.
\end{itemize}
This work opens up avenues for further investigation into Persian medical text analysis. The rest of this paper is organized as follows: Section \ref{sec:background} briefly reviews BERT-based language models in the medical domain. Then the procedure of pre-training \textsc{Sina-BERT} is presented in Section \ref{sec:approach}. After that, the evaluation results of \textsc{Sina-BERT} on downstream tasks are explained in Section \ref{sec:experiment}. Finally, concluding remarks are given in Section \ref{sec:conclusion}.

\section{Background}
\label{sec:background}
We have reviewed biomedical word embeddings and the BERT-based language models of medical-related domains below, as well as the pre-trained language models in Persian.

\subsection{Language Models for Medical Domain}
BioBERT \cite{lee2020biobert} is a domain-specific language model which was initialized by BERT and pre-trained on a large-scale biomedical corpus that contains PubMed abstracts and PubMed full-text articles. BioBERT is fine-tuned for three biomedical text mining tasks: NER, relation extraction, and question answering.

\textsc{SciBERT} \cite{beltagy2019scibert} is a language model which was pre-trained on a large multi-domain corpus of scientific publications. This corpus contains 1.14M papers randomly selected from Semantic Scholar. \textsc{SciBERT}  was evaluated on sequence tagging, sentence classification, and dependency parsing; all with data sets from a variety of scientific domains. 

Clinical BERT \cite{alsentzer-etal-2019-publicly} was pre-trained on a corpus of approximately 2 million clinical notes and improved the performance of clinical NLP tasks such as extracting Protected Health Information (PHI) during the process of anonymising medical records for de-identification. 

BEHR \cite{li2020behrt} is a Transformer-based deep neural sequence transduction model for electronic health records (EHR). The aim of training BEHR is to use a given patient’s past EHR to predict his/her future diagnoses (if any). This model was trained and evaluated on the data from nearly 1.6 million individuals.

MedBERT \cite{rasmy2020med} is another language model which was pre-trained on large-scale structured EHRs to benefit downstream disease-prediction tasks. This model was fine-tuned for the prediction of heart failure in patients with diabetes and the prediction of pancreatic cancer.

HQADeepHelper \cite{luo2020hqadeephelper} is a deep learning system that includes a wide range of healthcare question answering models; most of which are based on the pre-trained BERT or \textsc{SciBERT}. 

BioWordVec \cite{zhang2019biowordvec} is an open set of biomedical word embeddings that combines subword information from unlabeled biomedical text with a widely-used biomedical vocabulary. 


\subsection{Language Models in Persian}

There are two multi-lingual language models that support Persian: multi-lingual BERT \cite{devlin2018bert} and XLM-RoBERTa \cite{DBLP:conf/acl/ConneauKGCWGGOZ20} to the best of our knowledge. However, the size and the domain of the corpora used by them are not clear. 

ParsBERT \cite{farahani2020parsbert} is a mono-lingual BERT for the Persian language, which was pre-trained on a general domain corpus of 2.8M documents.  ParsBERT was evaluated on NER and sentiment analysis tasks. The domains of the data sets used in these evaluations are news and online shopping respectively.



\section{Approach}
\label{sec:approach}

Persian is among the under-resourced languages. Although there are language models that support Persian, none of them were pre-trained on a large Persian medical corpus. Understanding medical texts and solving medical tasks like question answering attract many researchers. However, the lack of a high-performance language model in this domain is a serious obstacle for them. In this section, we describe our Persian medical corpus and the details of pre-training \textsc{Sina-BERT}.

\subsection{Data Collection}


Although there are plenty of online Persian texts related to health and medicine, no large corpus is available. So, to train a medical language model in Persian, we had to gather together a large collection of texts from several online sources. The topic of these texts includes health, medicine, nursing, pharmacy, medical ethics and law, folk medicine, Persian medicine, lifestyle, nutrition, etc. This corpus contains 2.8M documents which were collected from the following sources: 
\begin{itemize}[noitemsep,nolistsep]
    \item health and medical news websites
    \item web sites publishing scientific materials about health, nutrition, lifestyle, etc.
    \item journals (abstract and full papers) and conference proceedings
    \item academic written materials
    \item medical reference books and theses
    \item online health-related forums
    \item medical and health-related pages of Instagram
    \item medical channels and groups of Telegram
\end{itemize}
The collected documents are then normalized and cleaned so they are free of HTML tags, hyperlinks, CSS, javascript, etc. 

Normalization is an essential pre-processing task in Persian because, unlike English, some Persian letters can be written in different forms with different ASCII codes. We have developed a new normalizer module in which mapping into a standard character is provided for all of the characters that appear in the corpus. Wired characters are mapped into empty characters, which means they are removed.

\subsection{Pre-Training \textsc{Sina-BERT}}

\textsc{Sina-BERT} is based on the BERT\textsubscript{BASE} model architecture \cite{devlin2018bert} which includes 12 hidden layers, 12 attention heads, and 768 hidden sizes. The total number of parameters of this configuration is 110M. The initialization of parameters is taken from ParsBERT \cite{farahani2020parsbert} which is a general domain BERT-base model in Persian. The tokenizer of ParsBERT is also borrowed. As per the original BERT and ParsBERT, the pre-training objective is the Masked Language Model (MLM), in which 15\% of tokens are randomly masked. The training batch size is 6, the learning rate is 5e-7, and each sequence contains 512 tokens at most.

\section{Validation on Medical Tasks}
\label{sec:experiment}

We validated \textsc{Sina-BERT} on four tasks. Since the lack of data sets for these tasks in Persian, we prepared annotated data for each task. These resources have been used in the evaluation of \textsc{Sina-BERT} and could be employed in further studies on Persian medical IR and QA tasks. In each task, \textsc{Sina-BERT}'s performance is compared with the below state-of-the-are language models already available in Persian: 
\begin{itemize}[noitemsep,nolistsep]
    \item BERT-Base, Multi-lingual Cased (mBERT) \cite{devlin2018bert} which is a multi-lingual language model that supports 102 languages including Persian.
    \item XLM-RoBERTa \cite{DBLP:conf/acl/ConneauKGCWGGOZ20} which is pre-trained for one hundred languages including Persian.
    \item ParsBERT \cite{farahani2020parsbert} which is the base model of \textsc{Sina-BERT}.
\end{itemize}
In contrast to \textsc{Sina-BERT}, the above language models were pre-trained on general domain data.

\subsection{Fill-in-the-Blank}
\label{sec:fill_in_the_blank}

The first task was fill-in-the-blank. We searched through a famous Persian website, Niniban\footnote{\url{http://niniban.com/}}, which is an online magazine. There are several forums on this site in which people discuss all medical and health-related matters, ask their questions and answer other people's questions. While the tone of the magazine's writing is completely formal, forums are mostly informal. Among all the materials on this website, 10,000 random sentences were selected. 15\% of the tokens in each sentence were then masked randomly. The Persian language model was used to predict the masked tokens. This data set was excluded from the corpus we used to pre-train \textsc{Sina-BERT}, so we considered the exact matching of the masked token with a predicted word to be true. Therefore, we consider the number of true cases divided by the total number of masked tokens to be an indication of the model's accuracy. Table \ref{tb:lm-mask-eval-acc} shows that \textsc{Sina-BERT} significantly outperforms other models. Also, ParsBERT is the second-best model because it was pre-trained with a larger Persian corpus in comparison with mBERT and XLM-RoBERTa. 

\begin{table}
\small
\centering
\caption{Accuracy of the Persian language models applied to the fill-in-the-blank task.}
\label{tb:lm-mask-eval-acc}
\begin{tabular}{lc}
    \toprule
    \textbf{Model} & \textbf{Accuracy}\\
    \midrule
    XLM-RoBERTa & 12.83 \\
    mBERT & 13.88 \\
    ParsBERT & 39.44 \\
    \textsc{Sina-BERT} & \textbf{50.71}  \\
    \bottomrule
\end{tabular}
\end{table}

\subsection{Medical Question Classification}
\label{sec:qc}

Medical Question Answering (MQA) systems have gained considerable attention. Question Classification (QC) is a major task within these systems because MQA systems may be designed to answer only some specific kinds of medical questions. Questions can be classified with consideration of  different aspects such as anatomy, disease causes, treatment type, etc. \cite{roberts2014automatically,roberts2016resource}. A common classification is based on the type of doctor that should respond to that question. 

To prepare a data set to validate \textsc{Sina-BERT} on this task, we used the QA data set we collected for the task of question retrieval, which will be explained in Section \ref{sec:retrieval}. Each QA has a meta-data that denotes the category of the question and the specialty of the doctors who answered that question. We selected ``pediatric Gastroenterology" as one of the most frequent categories in the database. Among the QA from this category, 1000 random samples were selected and labeled ``1". Also, 3400 random samples were selected from other categories and labeled ``0". To ensure that the automatic labels were correct, all samples were manually checked by two annotators. As a result, a data set containing 4400 QA was prepared. 

Using our data set, we ran a binary classifier. The [CLS] token of the last layer was fed into a linear classification layer. A dropout of 0.1 was applied and cross-entropy loss was optimized using Adam \cite{kingma2014adam}. The model was fine-tuned for 10 epochs using a batch size of 8 and a learning rate of 2e-5.

\begin{table}
\caption{Accuracy of the language models evaluated on the question classification task.}
\label{tb:binary-eval-acc}
\begin{adjustbox}{width=\columnwidth,center=\columnwidth}
\begin{tabular}{lcccc}
    \toprule
    \textbf{Model} & \textbf{Prec.} & \textbf{Rec.} & \textbf{Macro F$_1$} & \textbf{Accu.}\\
    \midrule
    mBERT & 88.41 & 87.41 & 87.89 & 90.80\\
    XLM-RoBERTa & 90.61 & 88.78 & 89.65 & 92.50 \\
    fastText + CNN & 90.90 & 91.30 & 91.10 & 93.52 \\
    ParsBERT & 93.01 & 93.13 & 93.07 & 94.66 \\
    \textsc{Sina-BERT} & \textbf{94.91} & \textbf{94.63} & \textbf{94.77} & \textbf{96.14} \\
    \bottomrule
\end{tabular}
\end{adjustbox}
\end{table}

Table \ref{tb:binary-eval-acc} shows the results of applying BERT-based models to the task of identifying questions related to pediatric gastroenterology. In addition to the BERT-based models, a Convolution Neural Network (CNN) was implemented which uses fastText \cite{bojanowski2017enriching} word embedding as the initialization of the Embedding layer. This embedding was trained on the same corpus that was used for the pre-training of \textsc{Sina-BERT}. According to the macro F$_1$ and accuracy measures, \textsc{Sina-BERT} outperforms other language models. The results obtained by this experiment confirm that \textsc{Sina-BERT} surpasses other Persian language models in understanding the content of medical questions.

\subsection{Medical Sentiment Analysis}
\label{sec:sentiment-data}
People often interact with other users with similar health conditions on social networks and share their experiences about doctors, drugs, treatments, or diagnosis. Therefore, sentiment analysis in medical setting \cite{yadav2018medical,denecke2015sentiment} has attracted much attention in recent years. 

To assess patients' satisfaction with their physician's performance, a data set containing 5,000 comments was collected from Persian online medical counseling websites. This data is mostly comments from people on the quality of the counsel they received from online doctors. They were manually labeled with Satisfaction, Un-satisfaction, and No-idea, so we defined a 3-classes classification task for this data set. From this set of comments, 5\% were used for testing, 10\% for validation, and the rest for the training.

To perform the evaluation, the embedding vectors of the comments generated by \textsc{Sina-BERT} and other base models were given to a CNN classifier. This classifier, which consists of 100 filters of different sizes [2, 3, 4, 5, 6] along with the max-pooling layer, predicts the label of each comment based on the given embedding vectors. The Adam optimizer with a learning rate of 2e-5 was used. The batch size was 16. The training was performed for 3 epochs.  

The results of the sentiment analysis based on \textsc{Sina-BERT} and other basic models are shown in Table \ref{tb:sentiment-eval-acc}. As can be seen, \textsc{Sina-BERT} has a better performance compared to multi-lingual models such as mBERT and XLM-RoBERTa. In the case of ParsBERT, its performance was close to \textsc{Sina-BERT} due to the fact that medical terminologies are normally less commonly used in the comments of users. For example, many people just said ``that was good", ``this doctor is not so good", ``last prescription didn't work for me at all", etc., which means most of the comments were short and simple and lacked professional vocabulary. 

\begin{table}
\caption{Accuracy of the language models evaluated on the medical sentiment analysis.}
\label{tb:sentiment-eval-acc}
\begin{adjustbox}{width=\columnwidth,center=\columnwidth}
\begin{tabular}{lcccc}
    \toprule
    \textbf{Model} & \textbf{Prec.} & \textbf{Rec.} & \textbf{Macro F$_1$} & \textbf{Accu.}\\
    \midrule
    mBERT & 0.91 & 0.90 & 90.06 & 0.90 \\
    fastText + CNN & 0.91 & 0.91 & 90.06 & 0.91 \\
    XLM-RoBERTa & 0.92 & 0.92 & 91.62 & 0.92 \\
    ParsBERT & 0.93 & 0.93 & 92.82 & 0.93 \\
    \textsc{Sina-BERT} & \textbf{0.95} & \textbf{0.94} & \textbf{94.49} & \textbf{0.94} \\
    \bottomrule
\end{tabular}
\end{adjustbox}
\end{table}

\subsection{Medical Question Retrieval}
\label{sec:retrieval}

A growing number of people including patients, doctors and healthcare professionals utilize Information Retrieval (IR) systems to seek answers to their questions. These questions vary from definitional questions, i.e., ``What is X?", to complex questions pertinent to a patient's illness such as how to assess symptoms in order to seek medical help and diagnosis \cite{cao2011askhermes}. 

In the task of question retrieval, a list of Question-Answer (QA) pairs are retrieved from a database of QA which are the most similar to the user's question. This retrieval system supports decision-making for diagnosis and treatment. We collected a set of 200K medical QA pairs. They were gathered from 20 Persian websites that provide online services for medical consultancy. Each question of this database has been already answered by at least one physician. These QA pairs are cleaned and normalized. Analysis of these medical questions shows that they vary in length from a short sentence to one or more paragraphs, as well as vary in tone from professional to personal and emotional. The average and standard deviation of question length are 69.0 and 78.3 tokens respectively. 

Most of the retrieval models take pre-trained representations and either 1) obtain a document representation from individual word representations that is subsequently used for ranking, or 2) combine representation similarities in some way to rank documents \cite{gysel2018neural}. 
A common method of generating question representation from word representation is to average the word representations. However, this basic representation can be improved. Therefore, we propose the following representations:
\begin{itemize}[noitemsep,nolistsep]
    \item \textsc{Sina-BERT}\_all: The average of all embeddings in the last layer of the network.
	 \item \textsc{Sina-BERT}\_rsw: It is similar to \textsc{Sina-BERT}\_all; but the stop-words are removed from the average pooling.   
    \item \textsc{Sina-BERT}\_kw: Instead of giving the complete question to the network, $n$ most important key phrases are selected together with two words before and after them as the context for key phrases. The enhanced key phrases are separated with [SEP] tokens, and this sequence is then given to the model. Key phrases are selected based on TF-IDF score. 
    \item \textsc{Sina-BERT}\_kw\_rcnt: It is similar to \textsc{Sina-BERT}\_kw; but two words after and before the key phrases are ignored just before the average pooling and only the embedding of key phrases are considered.
\end{itemize}
Therefore, we adopted an unsupervised approach toward ranking documents as follows: Given a user's query, the representation of this query is obtained and the similarity of it to all the questions of the database is calculated using their cosine similarity. The topmost similar ones are retrieved and presented to the user. In the next sections, we compare \textsc{Sina-BERT} with the current state-of-the-art models and report their scores.

\subsubsection{User-Oriented Evaluation}

In the first evaluation, 70 QA pairs of the database were selected randomly and separated from other QA pairs. These QA pair were supposed to be the user's queries that were given to the retrieval system. In response to each user's query, the most similar QA pair of the database (top one) was judged by a human. There is a multiple-choice format for the judgment: 
\begin{itemize}[noitemsep,nolistsep]
    \item Similar questions: two patients had similar conditions and their request was the same or very similar.
    \item Similar topics: two questions had similar topics; however they were not the same.
    \item Different topics: two questions had different topics.
\end{itemize}
These three options received scores of 1, 0.5, and 0 respectively. However, in a rigid evaluation, only the first case got a score of 1 and the others got 0. The accuracy was then calculated based on these scores. These human judgments were double checked to be fair across different language models. 

\begin{table}
\caption{Accuracy of the language models on the task of medical question retrieval evaluated by human judgment.}
\label{tb:manual-eval-acc}
\begin{adjustbox}{width=\columnwidth,center=\columnwidth}
\begin{tabular}{lcc}
    \toprule
    \textbf{Model} & \textbf{Accuracy (01)} & \textbf{Accuracy}\\
    \midrule
    XLM-RoBERTa & 18.57 & 30.00 \\
    mBERT & 18.57 & 31.42\\
    ParsBERT  & 25.71 & 32.86 \\
    \textsc{Sina-BERT}\_all & 30.00 & 41.43 \\
    \textsc{Sina-BERT}\_kw & 35.71 & 42.14 \\
    \textsc{Sina-BERT}\_rsw & 35.71 & 45.00 \\
    \textsc{Sina-BERT}\_kw\_rcnt & \textbf{35.71} & \textbf{47.85} \\
    \bottomrule
\end{tabular}
\end{adjustbox}
\end{table}

Table \ref{tb:manual-eval-acc} presents the scores obtained by \textsc{Sina-BERT} and other language models. All versions of \textsc{Sina-BERT} significantly outperform other BERT-based models. Comparing four methods for producing document representation from word embeddings reveals that extracting keywords and removing stop words improves the accuracy of \textsc{Sina-BERT}\_all by 1.7\% and 8.6\%, respectively. Adding the contexts into the keywords before feeding them into the model and removing the context word embeddings during the average pooling result the most improvement over \textsc{Sina-BERT}\_all, i.e. about 15.5\%, and result in the highest scores. This means that context words are necessary to produce the meaningful embedding of keywords; however, average of keyword's embedding is sufficient to build sentence representation.

\subsubsection{Paraphrased Test Data}
\label{sec:ir_paraph_data}

In the second evaluation, 200 QA pairs were selected from the database at random. These QAs were divided into four parts and each part was given to a human native in Persian with an academic degree to read the question carefully and produce a paraphrase for it. The guideline was to ``\textit{rewrite the question by changing the writing style, words, tone of the text, etc. at most at possible until no change in the meaning}''. 

Each paraphrased question was a query given to the retrieval system, and therefore the prime question is expected to be retrieved. To measure the performance of a retrieval system, we used the R@k metric, so we retrieved top k questions (k= 1, 5, and 10), and checked if the prime question was among the retrieved questions.

\begin{table}
\small
\centering
\caption{Performance of different methods of question representation on single-stage retrieval task using paraphrased test data.}
\label{tb:paraph-eval-acc}
\begin{adjustbox}{width=\columnwidth,center=\columnwidth}
\begin{tabular}{lccc}
    \toprule
    \textbf{Model} & \textbf{R@1} & \textbf{R@5} & \textbf{R@10}\\
    \midrule
    XLM-RoBERTa & 28.19 & 35.10 & 37.76 \\
    mBERT & 27.65 & 34.57 & 40.42 \\
    ParsBERT & 31.38 & 38.29 &  40.95  \\
    \textsc{Sina-BERT}\_all & 36.17 & 43.08 & 47.34 \\
    \textsc{Sina-BERT}\_kw & 40.42 & 45.74 & 48.93 \\
	\textsc{Sina-BERT}\_rsw & 42.02 & 50.00 & 54.78 \\    
    \textsc{Sina-BERT}\_kw\_rcnt & \textbf{44.14}  & \textbf{53.19} &  \textbf{55.31}\\
    
    \bottomrule
\end{tabular}
\end{adjustbox}
\end{table}

Table \ref{tb:paraph-eval-acc} presents the comparison of different language models. The overall scores are similar to Table \ref{tb:manual-eval-acc}, and \textsc{Sina-BERT}\_all outperforms all the state-of-the-art language models. Among the proposed methods for filtering tokens from the average pooling, \textsc{Sina-BERT}\_kw\_rcnt shows the most improvement of R@10; obtaining 16.8\% higher than the \textsc{Sina-BERT}\_{all}.

\begin{table}
\centering
\small
\caption{R@1 of the retrieval models applied to the noisy queries.}
\label{tb:noisy-eval-acc}

\begin{adjustbox}{width=\columnwidth}
\begin{tabular}{lccccc}
    \toprule
    \multirow{2}{*}{\textbf{Model}} & \multicolumn{5}{c}{\textbf{Noise Percentage}} \\
     & 0.1 & 0.2 & 0.3 & 0.4 & 0.5 \\
    \midrule
    XLM-RoBERTa & 86.1 & 25.0 & 4.9 & 0.9 & 0.3 \\
    mBERT & 97.8 & 83.2 & 34.6 & 9.3 & 1.3 \\
    ParsBERT & 98.4 & 79.8 & 28.8 & 6.9 & 1.3 \\
    \textsc{Sina-BERT}\_rsw & 48.1 & 32.1 & 13.5 & 4.3 & 0.8 \\
    \textsc{Sina-BERT}\_kw & 93.7 & 48.1 & 9.8 & 1.7 & 0.3 \\
    \textsc{Sina-BERT}\_all & \textbf{99.2} & 86.8 & 28.2 & 5.3 & 0.5 \\
    \textsc{Sina-BERT}\_kw\_rcnt & 99.1 & \textbf{97.1} & \textbf{85.1} & \textbf{57.9} & \textbf{23.5} \\
    \bottomrule
\end{tabular}
\end{adjustbox}
\end{table}

\subsubsection{Noisy Queries}
In the third evaluation, 1000 QAs from the database were selected randomly. In each question, $m$ percent of tokens were replaced with random tokens from the vocabulary. $m$ varies from 0.1, 0.2, to 0.5. This noisy data set is given to all methods. It is expected that the prime question is retrieved when the noisy question is the query. So, we evaluated the retrieval methods by using the R@1 metric. As Table \ref{tb:noisy-eval-acc} demonstrates, the highest scores are obtained by \textsc{Sina-BERT}\_kw\_{rcnt}. This method outperforms all systems by a substantial margin; especially for higher noise percentages.



\subsubsection{Comparing with Text Mining Methods}

In the last evaluation, different methods of document presentation were compared by using an unsupervised re-ranking approach: Firstly, an initial list of documents is retrieved by a simple and fast unsupervised bag-of-words method, e.g. BM25 \cite{robertson1995okapi}, which are then re-ranked by
the BERT-based models that produce the document representation from the word representation. 

In addition to \textsc{Sina-BERT}\_kw\_{rcnt}, we employed UKP-DistilBERT \cite{reimers-2020-multilingual-sentence-bert} and UKP-XLMR-paraph \cite{reimers-2020-multilingual-sentence-bert} which are two multi-lingual sentence embeddings. The training of these models is based on the idea that a translated sentence should be mapped to the same point in the vector space as the original sentence. Therefore a mono-lingual model, e.g. mBERT, is used to generate sentence embeddings for the source language and then train a new system on translated sentences to mimic the original model. These models are available in more than 50 languages including Persian. The similarity of two questions is computed based on the cosine similarity of sentence embedding of the two questions.


\begin{table}
\small
\centering
\caption{Comparison of different retrieval methods on the paraphrased data set.}
\label{tb:comaper-others}
\begin{tabular}{lcccc}
    \toprule
    \textbf{Model} & R@1 &  R@5 & R@10 & MRR\\
    \midrule
    UKP-DistilBERT & 36.36 & 49.73 & 54.54 & \\
	TF-IDF & 50.00 & 62.23 & 66.47 & \\    
    UKP-XLMR-paraph & 50.26 & 63.10 & 69.51 & \\
    \textsc{Sina-BERT} & \textbf{68.87} & \textbf{75.51} & \textbf{76.53}  & \\
    \bottomrule
\end{tabular}
\end{table}

The data set used in this experiment is the paraphrased data as was described in Section \ref{sec:ir_paraph_data}. Table \ref{tb:comaper-others} represents the recall scores obtained by different sentence representation methods. For a better comparison, we report the scores of the bag-of-word model of TF-IDF. The re-ranking method, which is based on \textsc{Sina-BERT}\_kw\_rcnt, significantly outperforms UKP-DistilBERT and UKP-XLMR-paraph. TF-IDF obtains a higher recall in comparison with the sentence representation method of UKP-DistilBERT. Although this is contrary to our expectations, the main reason for this result is that two paraphrased medical questions have many common keywords such as the names of drugs, names of diseases, names of medical treatment, etc. The medical domain is a named entity-rich area that changes the scores in the favor of the bag-of-word models when evaluated on the paraphrased test data.

Table \ref{tb:example} shows an example of a Persian medical query and the best-retrieved question by \textsc{Sina-BERT} in comparison with ParsBERT and UKP-Distill-BERT. This query is related to the field of pediatric gastroenterology, in which a mother asks for some advice for her baby who refuses to drink milk. All the retrieved questions also refer to the baby's nutrition. However, deeply considering the retrieved questions reveals that although there are many common words between them, the issue that araises only in the question on the left which was retrieved by \textsc{Sina-BERT}, is similar to the query and the others are relatively different. This example therefore shows that \textsc{Sina-BERT} can be employed in understanding medical documents such as field of pediatric gastroenterology.

\begin{table}
\centering
\caption{An example query with the best retrieved question. The last row shows the manual judgment.}
\label{tb:example}
\begin{tabular}{l}
    \resizebox{\columnwidth}{!}{
    \includegraphics{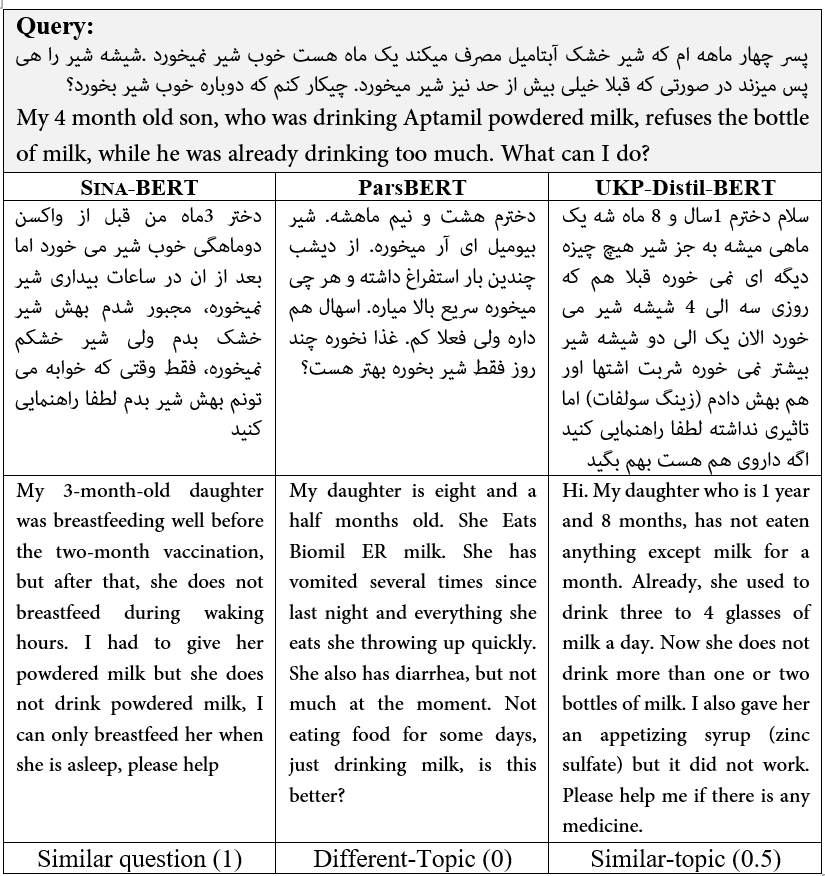}
}
\end{tabular}
\end{table}

\section{Conclusion and Future Work}
\label{sec:conclusion}

This paper was the first work on developing a medical language model in Persian. A BERT-based language model was pre-trained by collecting a large corpus of both formal and informal Persian texts from online resources. \textsc{Sina-BERT} was validated on four tasks and we have prepared a data set for each one. \textsc{Sina-BERT} outperformed the state-of-the-art Persian language models in all tasks. The margin between it and the other models in the task of question retrieval is much more than in the classification tasks of question classification and sentiment analysis; mainly because the supervision that exists in the classification tasks somewhat closes the gap between \textsc{Sina-BERT} and the other language models. However, for the unsupervised task of question retrieval, the significant differences reveal that training a language model across a large medical data set greatly benefits its understanding of related texts.

As for future works, there is a wide range of tasks in the area of Persian medical text analysis such as information extraction from clinical notes, medical NER, biological relation extraction, medical entity linking, disease prediction, etc., which can be solved using the \textsc{Sina-BERT}; subject to provision of the annotated data sets. Finally, the achievements of this research provide the foundation for further studies of Persian health and medical related tasks.

%
%
\bibliography{anthology,references}

\end{document}